\pdfoutput=1

\documentclass[11pt]{article}

\usepackage{EMNLP2023}

\usepackage{times}
\usepackage{latexsym}
\usepackage{float}

\usepackage[T1]{fontenc}

\usepackage[utf8]{inputenc}

\usepackage{microtype}

\usepackage{inconsolata}
\usepackage{graphicx}
\usepackage{listings}
\usepackage{dirtytalk}

\usepackage{xcolor}
\newcommand{\review}[1]{{\color{black}#1}}

\usepackage{{booktabs}}

\newcommand{\averitec}{AVeriTeC}

\makeatletter
\newcommand\footnoteref[1]{\protected@xdef\@thefnmark{\ref{#1}}\@footnotemark}
\makeatother

\title{AIC CTU system at \averitec{}: Re-framing automated fact-checking as a simple RAG task}

\author{Herbert Ullrich \\
AI Center @ CTU FEE\\
Charles Square 13\\
Prague, Czech Republic\\
\texttt{ullriher@fel.cvut.cz} \\\And
Tomáš Mlynář \\
AI Center @ CTU FEE\\
Charles Square 13\\
Prague, Czech Republic\\
\texttt{mlynatom@fel.cvut.cz} \\ \\\And
Jan Drchal \\
AI Center @ CTU FEE\\
Charles Square 13\\
Prague, Czech Republic\\
\texttt{drchajan@fel.cvut.cz} \\}

\begin{document}
\maketitle
\begin{abstract}
This paper describes our $3^{rd}$ place submission in the \averitec{} shared task in which we attempted to address the challenge of fact-checking with evidence retrieved in the wild using a simple scheme of Retrieval-Augmented Generation (RAG) designed for the task, leveraging the predictive power of Large Language Models.
We release our codebase\footnote{\url{https://github.com/aic-factcheck/aic_averitec}}, and explain its two modules -- the Retriever and the Evidence \& Label generator -- in detail, justifying their features such as MMR-reranking and Likert-scale confidence estimation.
We evaluate our solution on \averitec{} dev and test set and interpret the results, picking the GPT-4o as the most appropriate model for our pipeline at the time of our publication, with Llama 3.1 70B being a promising open-source alternative.
We perform an empirical error analysis to see that faults in our predictions often \review{coincide} with noise in the data or ambiguous fact-checks, provoking further research and data augmentation.

\end{abstract}


\section{Introduction}
\label{sec:introduction}
We release \review{a} pipeline for fact-checking claims using evidence retrieved from the web consisting of two modules -- a \textit{retriever}, which picks the most relevant sources among the available knowledge store\footnote{Due to the pre-retrieval step that was used to generate knowledge stores, our \say{retriever} module could more conventionally be referred to as a \say{reranker}, which we refrain from, to avoid confusion with reranking steps it uses as a subroutine.} and an \textit{evidence \& label generator} which generates evidence for the claim using these sources, as well as its veracity label. 

Our pipeline is a variant of the popular Retrieval-augmented Generation (RAG) scheme~\cite{rag}, making it easy to re-implement using established frameworks such as Langchain, Haystack, or our attached Python codebase for future research or to use as a baseline.

This paper describes our pipeline and the decisions taken at each module, achieving a simple yet efficient RAG scheme that improves dramatically across the board over the baseline system from~\cite{averitec2024}, and scores third in the \averitec{} leaderboard as of August 2024, with an \averitec{} test set score of 50.4\%.

\begin{figure}[h]
    \centering
    \includegraphics[width=0.47\textwidth]{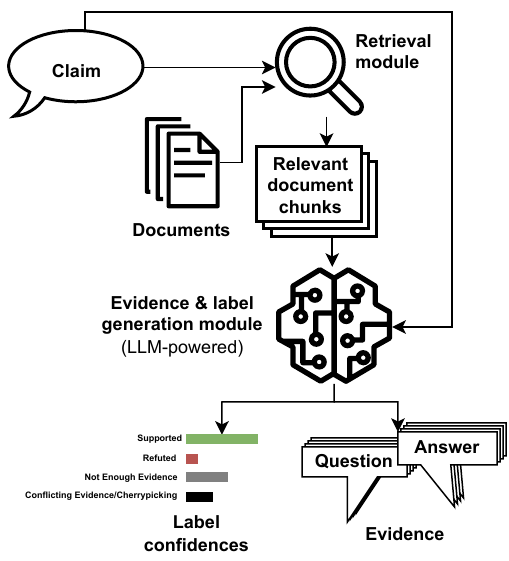}
    \caption{Our pipeline}
    \label{fig:pipeline}
\end{figure}

\section{Related work}
\label{sec:relwork}
\label{avscore}
\begin{enumerate}
    \item \textbf{\averitec{} shared task}~\cite{averitec2024} releases the dataset of real-world fact-checked claims, annotated with evidence available at the date the claim was made.
    
    \review{It proposes the \textbf{\averitec{} Score} -- a method of unsupervised scoring of fact-checking pipeline against this gold data using Hungarian METEOR score, matching the evidence questions (Q) or the whole evidence (Q+A).
    The score is then calculated as the proportion of claims with accurate label and sound evidence (using a threshold for Hu-METEOR such as 0.25) among all claims in the dataset, giving an estimate on \say{how often the whole fact-checking pipeline succeeds end to end}.}

    The provided \textbf{baseline} is a pipeline of search query generation, API search (producing a knowledge store), sentence retrieval, Question-and-answer (QA) generation, QA reranking, QA-wise claim classification and label aggregation, achieving an overall \averitec{} test set score of 11\%.  
    \item \review{\textbf{FEVER Shared Task}~\cite{thorne-etal-2018-fact}, a predecessor to the \averitec{}, worked with a similar dataset engineered on top of the enclosed domain Wikipedic data rather than real-world fact-checks. 
    Its top-ranking solutions used a simpler pipeline of Document Retrieval, Sentence Reranking and Natural Language Inference, improving its modules in a decoupled manner and scoring well above 60\% in a similarly computed FEVER score~\cite{thorne-etal-2018-fever} on this data.}
    \item \textbf{Our previous research} on fact-checking pipelines~\cite{Ullrich2023,drchal2023pipelinedatasetgenerationautomated} using data similar to FEVER and \averitec{} shows significant superiority of fact-checking pipelines that \textbf{retrieve the whole documents} for the inference step, rather than retrieving out-of-context sentences.
    \item \textbf{Retrieval-Augmented Generation (RAG) for Knowledge-Intensive Tasks}~\cite{rag} takes this a step further, leveraging Large Language Model (LLM) for the task, providing it the whole text of retrieved documents (each a chunk of Wikipedia) and simply instructing it to predict the evidence and label on top of it, achieving results within 4.3\% from the FEVER state of the art by the time of its publication (December 2020) \textit{without} engineering a custom pipeline for the task.
\end{enumerate}


\section{System description}
\label{sec:system}
\review{Our system design prioritizes simplicity, and its core idea is using a straightforward RAG pipeline without engineering extra steps, customizing only the retrieval step and LLM prompting } (Listing~\ref{lst:llm_system_prompt} \review{in Appendix~\ref{appendix_sec:system_prompt}}).
Despite that, this section describes and justifies our decisions taken at each step, our additions to the naive version of RAG modules to tune them for the specific task of fact-checking, and their impact on the system performance.

\subsection{Retrieval module}
\label{retrieval}
To ease comparison with the baseline and other systems designed for the task, our system does not use direct internet/search-engine access for its retrieval, but an \averitec{} \textit{knowledge store} provided alongside each claim.

\review{To use our pipeline in the wild, our retrieval module is decoupled from the rest of the pipeline and can be swapped out in favour of an internet search module such as SerpApi\footnote{\url{https://serpapi.com/}} as a whole, or it can be used on top of a knowledge store emulated using large crawled corpora such as CommonCrawl\footnote{\url{https://commoncrawl.org/}} and a pre-retrieval module.}

\subsubsection{Knowledge stores}
Each claim's knowledge store contains pre-scraped results for various search queries that can be derived from the claim using human annotation or generative models.
The knowledge stores used with ours as well as the baseline system can be downloaded from the \averitec{}  dataset page\footnote{\url{https://fever.ai/dataset/averitec.html}}, containing about 1000 pre-scraped \textit{documents}\footnote{\label{devsetnote}The numbers are orientational and were computed on knowledge stores provided for the \averitec{}  dev set.}, each consisting of $28$ sentences at median\footnoteref{devsetnote}, albeit varying wildly between documents.
The methods used for generating the knowledge stores are explained in more detail by~\citet{averitec2024}.

Our retrieval module then focuses on picking a set of $k$ ($k=10$ in the examples below, as well as in our submitted system) most appropriate document chunks to fact-check the provided claim within this knowledge store.

\subsubsection{Angle-optimized embedding search}
\label{sec:knn}
Despite each article in any knowledge store only needing to be compared \textit{once} with its \textit{one specific} claim, which should be the use-case for CrossEncoder reranking~\cite{dejean2024thoroughcomparisoncrossencodersllms}, our empirical preliminary experiments made us favour a \textit{cosine-similarity} search based on vector embeddings instead.
It takes less time to embed the whole knowledge store into vectors than to match each document against a claim using crossencoder, and the produced embeddings can be re-used across experiments.

For our proof of concept, we explore the MTEB~\cite{muennighoff-etal-2023-mteb} benchmark leaderboard, looking for a reasonably-sized open-source embedding model, ultimately picking Mixedbread's mxbai-large-v1~\cite{li-li-2024-aoe,emb2024mxbai} optimized for the cosine objective fitting our inteded use.

\review{To reduce querying time at a reasonable exactness tradeoff, we use Faiss index~\cite{douze2024faiss,johnson2019billion} to store our vectors, allowing us to only precompute semantical representation once, making the retriever respond rapidly in empirical experiments, allowing a very agile prototyping of novel methods to be used.}

\subsubsection{Chunking with added context}
Our initial experiments with the whole \averitec{}  documents for the Document Retrieval step have revealed a significant weakness -- while most documents fit within the input size of the embedding model, outliers are common, often with \textit{hundreds of thousands} characters, exceeding the 512 input tokens with little to no coverage of their content.

Upon further examination, these are typically PDF documents of legislature, documentation and communication transcription -- highly relevant sources real fact-checker would scroll through to find the relevant part to refer. 

This workflow inspires the use of \textit{document chunk retrieval} as used in~\cite{rag}, commonly paired with RAG.
We partition each document into sets of its sentences of combined length of $N$ characters at most.
To take advantage of the full input size of the vector embedding model we use for semantical search, we \review{arbitrarily} set our bound $N=512*4=2048$, \review{where 512 is} the input dimension of common embedding models, 4 often being used as a rule-of-thumb number of characters per token for US English in modern tokenizers~\cite{tokens}.

Importantly, each chunk is  assigned metadata -- the source URL, as well as the full text of the next and previous chunk within the same document.
This way, chunks can be presented to the LLM along with their original context in the generation module, where the length constraint is much less of an issue than in vector embedding.
As shown in~\cite{drchal2023pipelinedatasetgenerationautomated}, fact-checking models benefit from being exposed to larger pieces of text such as paragraphs or entire documents rather than out-of-context sentences.
Splitting our data into the maximum chunks that fit our retrieval model and providing them with additional context may help down the line, preventing the RAG sources from being semantically incomplete.

\subsubsection{Pruning the chunks}
While the chunking of long articles prevents their information from getting lost to retriever, it makes its search domain too large to embed on demand.
As each of the thousands of claims has its own knowledge store, each of possibly tens of thousands of chunks, we seek to omit the chunks having little to no common tokens with our claim using an efficient BM25~\cite{bm25} search for the nearest $\omega$ chunks, setting the $\omega$ to 6000 for dev and 2000 for test claims. 
This yields a reasonably-sized document store for embedding each chunk into a vector, taking an average of 40 s to compute and store using the method described in Section~\ref{sec:knn} for each dev-claim using our Tesla V100 GPU.

This allows a quick and agile production of vectorstores for further querying and experimentation, motivated by the \averitec{}  test data being published just several days before the announced submission deadline.
\review{The pruning also keeps the resource intensity moderate for real-world applications.
However, if time is not of the essence, the step can be omitted.}

\subsubsection{Diversifying sources: MMR}
\review{Our choice of embedding search based on the entire claim rather than generating \say{search queries} introduces less noise and captures the semantics of the whole claim.
It is, however, prone to redundancy among search results, which we address using a reranking by the results' Maximal Marginal Relevance (MMR)~\cite{carbonell-mmr}, a metric popular for the RAG task, which maximizes the search results' score computed as (for $D_i\in P$)
$$\lambda \cdot \mathrm{Sim}(D_i, Q) - (1-\lambda) \cdot \max_{D_j \in S} \mathrm{Sim}(D_i, D_j)$$
$Sim$ denoting the cosine-similarity between embeddings, $Q$ being the search query, and $P$ the pre-fetched set of documents (by a search which simply maximizes their $Sim$ to $Q$), forming $S$ as the final search result, by adding each $D_i$ as MMR-argmax one by one, until reaching its desired size.}

In our system, we set $\lambda=0.75$ to favour relevancy rather than diversity, $|S|=10$ and $|P| = 40$, obtaining a set of diverse sources relevant to each claim at a fraction of cost and complexity of a query-generation driven retrieval, such as that used in~\cite{averitec2024}.

\subsection{Evidence \& label generator}
\label{sec:generation}
The second and the last module on our proposed pipeline for automated fact checking is the Evidence \& Label Generator, which receives a claim and $k$ sources (document chunks), and returns $l$ (in our case, $l=10$) question-answer pairs of evidence abstracted from the sources, along with the veracity verdict -- in \averitec{} dataset, a claim may be classified as \textit{Supported}, \textit{Refuted}, \textit{Not Enough Evidence}, or \textit{Conflicting Evidence/Cherrypicking} with respect to its evidence.

Our approach leverages a Large Language Model (LLM), instructing it to output both evidence and the label in a single step, as a chain of thought.
We rely on JSON-structured output generation with source referencing using a numeric identifier, we estimate the label confidences using Likert-scale ratings.
The full system prompt can be examined in Listing~\ref{lst:llm_system_prompt} \review{in Appendix~\ref{appendix_sec:system_prompt},} and this section further explains the choices behind it.

\subsubsection{JSON generation}

To be able to collect LLM's results programmatically, we exploit their capability to produce structured outputs, which is on \review{the} rise, with datasets available for tuning~\cite{tang2024strucbenchlargelanguagemodels} and by the time of writing of this paper (August 2024), systems for strictly structured prediction are beginning to be launched by major providers~\cite{json}.

Despite not having access to such structured-prediction API by the time of \averitec{} shared task, the current generation of models examined for the task (section~\ref{sec:chosen_llms}) rarely strays from the desired format if properly explained within a system prompt -- we instruct our models to output a JSON of pre-defined properties (see prompt Listing~\ref{lst:llm_system_prompt} \review{in Appendix~\ref{appendix_sec:system_prompt}}) featuring both evidence and the veracity verdict for a given claims.

Although we implement fallbacks, less than 0.5\% of our predictions \review{threw} a parsing exception throughout experimentation, and could be easily recovered using the same prompting again, exploiting the intrinsic randomness of LLM predictions.

\subsubsection{Chain-of-thought prompting}
While JSON dictionary should be order-invariant, we can actually exploit the order of outputs in our output structure to make LLMS like GPT-4o output better results~\cite{cot}.
This is commonly referred to as the \say{chain-of-thought} prompting -- if we instruct the autoregressive LLM to first output the evidence (question, then answer), then a set of all labels with their confidence ratings (see section~\ref{likert}) and only then the final verdict, its prediction is both cheaper as opposed to implementing an extra module, as well as more reliable, as it must attend to all of the intermediate steps as well.

\subsubsection{Source referring}
To be able to backtrack the generated evidence to the urls of the used sources, we simply augment each question-answer pair with a source field.
We assign a 1-based index\footnote{\review{We chose the 1-based source indexing to exploit the source-referring data in LLM train set such as Wikipedia, where source numbers start with 1. The improvement in quality over 0-based indexing was not experimentally tested.}}  to each of the sources to facilitate tokenization and prompt the LLM to refer it as the source ID with each evidence it generates.
While hallucination can not be fully prevented, it is less common than it may appear -- with RAG gaining popularity, the models are being trained to cite their sources using special citation tokens~\cite{menick2022teachinglanguagemodelssupport}, not dissimilarly to our proposal.

\subsubsection{Dynamic few-shot learning}
To utilise the few-shot learning framework~\cite{fewshot} shown to increase quality of model output, we provide our LLMs with examples of what we expect the model to do.
To obtain such examples, our evidence generator looks up the \averitec{} train set using BM25 to get the 10 most similar claims, providing them as the few-shot examples, along their gold evidence and veracity verdicts.
Experimentally, we also few-shot our models to output an \textit{answer type} (\textit{Extractive}, \textit{Abstractive}, \textit{Boolean},\dots) as the \textit{answer type} is listed with each sample anyways, and we have observed its integration into the generation task to slightly boost our model performance.

\subsubsection{Likert-scale label confidences}
\label{likert}
Despite modern LLMs being well capable of predicting the label in a \say{pick one} fashion, research applications such as ours may prefer them to output a probability distribution over all labels for two reasons.

Firstly, it measures the confidence in each label, pinpointing the edge-cases, secondly, it allows ensembling the LLM classification with any other model, such as Encoders with classification head finetuned on the task of Natural Language Inference (NLI) (see section~\ref{subsubsec:ensembling}).

As the LLMs and other token prediction schemes struggle with the prediction of continuous numbers which are notoriously hard to tokenize appropriately~\cite{golkar2023xvalcontinuousnumberencoding}, we come up with a simple alternative: instructing the model to print each of the 4 possible labels, along with their Likert-scale rating: 1 for \say{strongly disagree}, 2 for \say{disagree}, 3 for \say{neutral}, 4 for \say{agree} and 5 for \say{strongly agree}~\cite{likert1932technique}.

On top of the ease of tokenization, Likert scale's popularity in psychology and other fields such as software testing~\cite{likertstudy} adds another benefit -- both the scale itself and its appropriate usage were likely demonstrated many times to LLMs during their unsupervised training phase.

To convert the ratings such as \texttt{\{\say{Supported}:2, \say{Refuted}:5, \say{Cherrypicking}:4, \say{NEE}:2\}} to a probability distribution, we simply use softmax~\cite{NIPS1989_0336dcba}.
While the label probabilities are only emulated (and may only take a limited, discrete set of values) and the system may produce ties, it gets the job done until further research is carried out.

\subsubsection{Choosing LLM}
\label{sec:chosen_llms}
In our experiments, we have tested the full set of techniques introduced in this section, computing the text completion requests with:
\begin{enumerate}
    \item GPT-4o (version \texttt{2024-05-13})
    \item Claude-3.5-Sonnet (\texttt{2024-06-20}), using the Google's Vertex API
    \item LLaMA 3.1 70B, in the final experimets to see if the pipeline can be re-produced using open-source models
\end{enumerate} 

Their comparison can be seen in tables~\ref{tab:nli} and~\ref{tab:pipeline_scores}; for our submission in the \averitec{}  shared task, GPT-4o was used.
\section{Other examined approaches}
\label{sec:failed}
In this section, we also describe a third, optional module we call the \textit{veracity classifier}, which takes the claim and its evidence generated by our evidence \& label generator~(section~\ref{sec:generation}) and predicts the veracity label independently, based on the suggested evidence, using a fine-tuned NLI model.
We also describe the options of its ensembling with veracity labels predicted in the generative step (section~\ref{likert}).

The absence of a dedicated veracity classifier has not been shown to decrease the performance of our pipeline significantly (as shown, e.g., in tables~\ref{tab:pipeline_scores} and~\ref{tab:nli}) so we suggest to omit this step altogether and we proceed to participate in the \averitec{}  shared task without it, proposing a clean and simple RAG pipeline without the extra step (Figure~\ref{fig:pipeline}) for the fact-checking task.

\subsection{Single-evidence classification with label aggregation}
In the earliest stages of experimenting, we utilized the baseline classifier provided by \averitec{} authors\footnote{\url{https://huggingface.co/chenxwh/AVeriTeC}}~\cite{averitec2024}.
It is based on the BERT~\cite{devlin-etal-2019-bert} and was further fine-tuned on the \averitec{}  dataset~\cite{averitec2024}. 
It takes one claim and one question-answer evidence as input -- each claim therefore has multiple classifications, one for each evidence. The classifications are then aggregated using a heuristic of several if-clauses to determine the final label. 

We experiment with altering this heuristic (e.g. by making \textit{not enough evidence} the final label only when no other labels are present at any evidence), and training NLI models that could work better with it, such as 3-way DeBERTaV3~\cite{he2023debertav3improvingdebertausing} without a breakthrough result, motivating a radically different approach.

\subsection{Multi-evidence classification}
\label{subsubsec:concatenation}
The multi-evidence approach is to fine-tune a 4-way Natural Language Inference (NLI) classifier, using the full scope of evidence directly at once, without heuristics.
For that, we concatenate all of the evidence together using a separator \texttt{[SEP]} token. This allows the model to know exact question-answer borders, albeit using a space has turned out to be just as accurate as the experiments went on. As the veracity verdict should be independent of the evidence ordering, we also experiment with sampling different permutations in the fine-tuning step to increase the size of our data.

We carry out the fine-tuning using the \averitec{} train split with gold evidence and labels on \mbox{DeBERTaV3}~\cite{he2023debertav3improvingdebertausing} in two variants: the original large one\footnote{\url{https://huggingface.co/microsoft/deberta-v3-large}} and one pre-finetuned on NLI tasks\footnote{\url{https://huggingface.co/cross-encoder/nli-deberta-v3-large}}, and also Mistral-7B-v0.3 model\footnote{\url{https://huggingface.co/mistralai/Mistral-7B-v0.3}} with a classification head (MistralForSequenceClassification) provided by the Huggingface Transformers library~\cite{wolf-etal-2020-transformers} that utilizes the last token. In the preliminary testing phase, the original DeBERTaV3 Large performed the best and was used in all other experimental settings.

From the approaches described above, we achieved the best results for the development split with gold evidence and labels with a model without permuting the evidence, achieving 0.71 macro $F_1$ score using a space-separation. The \texttt{[SEP]} model achieved a comparable 0.70 macro $F_1$ score, and the random order model performed worse with a 0.67 macro $F_1$ score, all improving significantly upon baseline, yet falling behind the capabilities of generating the labels alongside evidence in a single chain-of-thought. 
We provide our best DeBERTaV3 finetuned model publicly in a Huggingface repository\footnote{\url{https://huggingface.co/ctu-aic/deberta-v3-large-AVeriTeC-nli}}.

\subsection{Ensembling classifiers}
\label{subsubsec:ensembling}

Encouraged by the promising results of our multi-evidence classifiers, we go on to try to ensemble the models with LLM predictions from section~\ref{likert}, using a weighted average of the class probabilities of our models.
We have experimented with multiple weight settings: 0.5:0.5 for even votes, 0.3:0.7 in favour of the LLM to exploit its accuracy while tipping its scales in cases of a more spread-out label probability distribution, as well as 0.1:0.9 to use the fine-tuned classifier only for tie-breaking, listing the results in Table~\ref{tab:nli}.

We also tried tuning our ensemble weights based on a subset of the dev split, without a breakthrough in accuracy on the rest of dev samples.

The last method we tried was stacking using logistic regression. However, this setup classified no labels from \textit{Not Enough Evidence} and \textit{Conflicting Evidence/Cherrypicking}, and we could not achieve reasonable results. For logistic regression, we used the scikit-learn library~\cite{scikit-learn}.

We conclude that the augmentation of the pipeline from Figure~\ref{fig:pipeline} with a classification module using a single NLI model or an ensemble with LLM is unneccessary, as it adds complexity and computational cost without paying off on the full pipeline performance (Table~\ref{tab:pipeline_scores}).

\subsection{Conflicting Evidence/Cherrypicking detection}

During the experiments, we discovered that classifying the \textit{Conflicting Evidence/Cherrypicking} class is the most challenging task, achieving a near-zero $F_1$-score across our various prototype pipelines.
To overcome this problem, we tried to build a binary classifier with cherrypicking as positive class. We tried to use the DeBERTaV3 Large model with both basic and weighted cross-entropy loss (other experimental settings were the same as in section~\ref{subsubsec:concatenation}), but it could not pick up the training task due to the \textit{Conflicting Evidence/Cherrypicking} underrepresentation in train set -- less than 7\% of the samples carry the label. 

Even after exploring various other methods, we did not get a reliable detection scheme for this task, perhaps motivating a future collection of data that represents the class better.
While writing this system description paper, we found an interesting research by~\citet{jaradat2024contextawaredetectioncherrypickingnews} that uses a radically different approach to detect cherrypicking in newspaper articles.


\section{Results and analysis}
\label{sec:results}

We examine our pipeline results using two sets of metrics -- firstly, we measure the prediction accuracy and $F_1$ over predict labels without any ablation, that is obtaining predicted labels using the predicted evidence generated on top the predicted retrieval results. 
While the retrieval module is fixed throughout the experiment (a full scheme described in section~\ref{retrieval}), various Evidence \& Label generators and classifiers are compared in Table~\ref{tab:nli}, showcasing their performance on the same sources.
The results show that if we disregard the quality of evidence, models are more or less interchangeable, without a clear winner across the board -- an ensemble of DeBERTA and Claude-3.5-Sonnet gives the best $F_1$ score, while GPT-4o scores 72\% accuracy.
\begin{table}    \centering
    \setlength\tabcolsep{3pt} 
    \resizebox{\columnwidth}{!}{%
    \begin{tabular}{lcccc}
        \toprule
        \textbf{Classifier} & \textbf{Acc} & \textbf{$F_1$} & \textbf{Prec.} & \textbf{{\footnotesize {Recall}}} \\
        \midrule
        GPT4o & 0.72 & 0.46 & 0.48 & 0.47 \\
        Claude 3.5 Sonnet & 0.64 & 0.49 & 0.50 & 0.52 \\
        DeBERTa & 0.63 & 0.39 & 0.40 & 0.41 \\
        DeBERTa - random@10 & 0.65 & 0.41 & 0.41 & 0.44 \\
        $0.5\cdot\mbox{DeBERTa}+0.5\cdot\mbox{GPT4o}$ & 0.70 & 0.43 & 0.41 & 0.45 \\
        $0.5\cdot\mbox{DeBERTa}+0.5\cdot\mbox{Claude}$ & 0.68 & 0.47 & 0.50 & 0.49 \\
        $0.3\cdot\mbox{DeBERTa}+0.7\cdot\mbox{GPT4o}$ & 0.72 & 0.45 & 0.45 & 0.46 \\
        $0.3\cdot\mbox{DeBERTa}+0.7\cdot\mbox{Claude}$ & 0.66 & \textbf{0.50} & \textbf{0.51} & \textbf{0.53} \\
        $0.1\cdot\mbox{DeBERTa}+0.9\cdot\mbox{GPT4o}$ & {0.72} & 0.39 & 0.46 & 0.43 \\
        $0.1\cdot\mbox{DeBERTa}+0.9\cdot\mbox{Claude}$ & 0.64 & 0.49 & 0.50 & 0.54 \\
        \midrule
        Llama 3.1 & \textbf{0.73} & 0.44 & 0.43 & 0.46 \\
        \bottomrule
    \end{tabular}
    }
    \caption{Evalution of the label generators, classifier models and their ensembles on the \averitec development set. $F_1$, Precision and Recall are computed as macro-averages. The random@10 suffix indicates that the classifier used average of 10 different random orders of QA pairs for each claim. GPT4o stands for the Likert classifier based on GPT-4o, Claude 3.5 Sonnet is the Likert classifier based on Claude 3.5 Sonnet, and DeBERTa is classifier based on DeBERTaV3 Large fine-tuned on \averitec{} gold evidence and labels.}
    \label{tab:nli}
\end{table}
\begin{table*}    \centering
    \begin{tabular}{l | c c c | c c c}
    \hline
    &\multicolumn{3}{c|}{\textbf{Dev Set Scores}} & \multicolumn{3}{c}{\textbf{Test Set Scores}}  \\
    \textbf{Pipeline Name} & \textbf{Q only} & \textbf{Q+A} & \textbf{\averitec{}} & \textbf{Q only} & \textbf{Q+A} & \textbf{AVeriTeC} \\ \hline
    \textbf{GPT-4o (full-featured pipeline)}      & \textbf{0.46} & \textbf{0.29} & \textbf{0.42} & 0.46 & \textbf{0.32} & \textbf{0.50}\\
    GPT-4o (simplified pipeline)         & 0.45 & 0.28 & 0.38 & 0.45 & 0.30 & 0.47 \\
    Claude-3.5-Sonnet (full-featured)             & 0.43 & 0.28 & 0.35 & 0.42 & 0.30 & 0.46 \\
    GPT-4o (with DeBERTa classification)              & 0.45 & 0.28 & 0.36 & -- & -- & --\\
    \averitec{} baseline            & 0.24 & 0.19 & 0.09 & 0.24 & 0.20 & 0.11\\
    \hline
    Llama 3.1 70B (full-featured) & \textbf{0.46} & 0.27 & 0.36 & \textbf{0.47} & 0.29 & 0.42\\
    \bottomrule
    \end{tabular}
    \caption{Comparison of Pipeline Scores on Dev and Test Sets. \review{Q, Q+A are Hu-METEOR scores against gold data,} AVeriTeC scores \review{are calculated as referred in section~\ref{avscore} thresholded at 0.25}. \say{Full-featured} pipelines use the all the improvement techniques introduced in section~\ref{sec:system}, while the simplified pipeline omits the dynamic few-shot learning, answer-type-tuning and Likert-scale confidence emulation described in section~\ref{sec:generation}}
    \label{tab:pipeline_scores}\end{table*}

In real world, however, the evidence quality is critical for the fact-checking task.
We therefore proceed to estimate it using the hu-METEOR evidence question score, QA score and \averitec{} score benchmarks briefly explained in Section~\ref{avscore} and in greater detail in~\cite{averitec2024}.
We use the provided \averitec{} scoring script to calculate the values for Table~\ref{tab:pipeline_scores}, using its EvalAI blackbox to obtain the test scores without seeing the gold test data.

The latter experiments shown in Table~\ref{tab:pipeline_scores} suggests the superiority of GPT-4o to predict the results for our pipeline with a margin.
Even if we simplify the evidence \& label generation step by omitting the dynamic few-shot learning (section~\ref{sec:generation}), answer-type tuning and Likert-scale confidence emulation, it still scores above others, also showing that our pipeline can be further simplified when needed.
Regardless of the LLM in use, the results of our pipeline improve upon the \averitec{} baseline dramatically.

Posterior to the original experiments and to the \averitec{} submission deadline, we also compute the pipeline results using an open-source model -- the Llama 3.1 70B\footnote{\url{https://huggingface.co/hugging-quants/Meta-Llama-3.1-70B-Instruct-AWQ-INT4}}~\cite{dubey2024llama3herdmodels} obtaining encouraging scores, signifying our pipeline being adaptable to work well without the need to use a blackboxed proprietary LLM.

\subsection{API costs}
During our experimentation July 2024, we have made around 9000 requests to OpenAI's \texttt{gpt-4o-2024-05-13} batch API, at a total cost of \$363.
This gives a mean cost estimate of \$0.04 per a single fact-check (or \$0.08 using the API without the batch discount) that can be further reduced using cheaper models, such as \texttt{gpt-4o-2024-08-06}.

We argue that such costs make our model suitable for further experiments alongside human fact-checkers\review{,} whose time spent reading through each source and proposing each evidence by themselves would certainly come at a higher price.

Our successive experiments with Llama 3.1~\cite{dubey2024llama3herdmodels} show promising results as well, nearly achieving parity with GPT.
The use of open-source models such as LLaMa or Mistral allows running our pipeline on premise, without leaking data to a third party and billing anything else than the computational resources.
For further experiments, we are looking to integrate them into the attached Python library using VLLM~\cite{vllm}.

\subsection{Error analysis}
In this section, we provide the results of an explorative analysis of 20 randomly selected samples from the development set. We divide our description of the analysis into the pipeline and dataset errors.

\subsubsection{Pipeline errors}
Our pipeline tends to rely on unofficial (often newspaper) sources rather than official government sources, e.g., with a domain ending or containing \texttt{gov}. On the other hand, it seems that the annotators prefer those sources. This could be remedied by implementing a different source selection strategy, preferring those official sources. For an example, see Listing~\ref{lst:gov_error} \review{in Appendix~\ref{appendix_sec:errors}}.

Another thing that could be recognised as an error is that our pipeline usually generates all ten allowed questions (upper bound given by the task~\cite{averitec2024}). The analysis of the samples shows that the last questions are often unrelated or redundant to the claim and do not contribute directly to better veracity evaluation. However, since the classification step of our pipeline is not dependent on the number of question-answer pairs, this is not a critical error.
Listing~\ref{lst:unrelated_questions} \review{in Appendix~\ref{appendix_sec:errors}} shows an example of a \review{data point} with some unrelated questions.

When the pipeline generates extractive answers, it sometimes happens that the answer is not precisely extracted from the source text but slightly modified. An example of this error can be seen in Listing~\ref{lst:extractive_error} \review{in Appendix~\ref{appendix_sec:errors}}. This error is not critical, but it could be improved in future works, e.g. using post-processing via string matching.

Individual errors were also caused by the fact that we do not use the claim date in our pipeline and because our pipeline cannot analyse PDFs with tables properly. The last erroneous behaviour we have noticed is that the majority of questions and answers are often generated from a single source. This should not be viewed as an error, but by introducing diversity into the sources, the pipeline would be more reliable when deployed in real-world scenarios.

\subsubsection{Dataset errors}
During the error analysis of our pipeline, we also found some errors in the \averitec{} dataset that we would like to mention. In some cases, there is a leakage of PolitiFact \review{or Factcheck.org} fact-checking articles where the claim is already fact-checked. This leads to a situation where our pipeline gives a correct verdict using the leaked evidence. However, annotators gave a different label (often Not Enough Evidence). \review{An example of this error is shown in Listing~\ref{lst:polifact_leakage} in Appendix~\ref{appendix_sec:errors}}. 

Another issue we have noticed is the inconsistency in the questions and answers given by annotators. \review{Sometimes, they tend to be longer, including non-relevant information, while some are much shorter, as seen in Listing~\ref{lst:different_lengths} in Appendix~\ref{appendix_sec:errors}}. The questions are often too general, or the annotators seem to use outside knowledge. This inconsistency in the dataset leads to a decreased performance of any models evaluated on this dataset.

\subsubsection{Summary}
Despite the abovementioned errors, the explorative analysis revealed that our pipeline consistently gives reasonable questions and answers for the claims. Most misclassified samples in those 20 data points were due to dataset errors.


\section{Conclusion}
\label{sec:conclusion}
In this paper, we describe the use and development of a RAG pipeline over real world claims and data scraped from the web for the \averitec{} shared task.
Its main advantage are its simplicity, consisting of just two decoupled modules -- Retriever and an Evidence~\& Label Generator -- and leveraging the trainable parameters of a LLM rather than on complex pipeline engineering.
The LLMs capabilities may further improve in future, making the upgrades of our system trivial.

In section~\ref{sec:system}, we describe the process of adding features to both modules well in an iterative fashion, describing real problems we have encountered and the justifications of their solution, hoping to share our experience on how to make such systems robust and \review{well-performing}.
We publish our failed approaches in section~\ref{sec:failed} and the metrics we observed to benchmark our systems in section~\ref{sec:results}. 
We release our Python codebase to facilitate further research and applications of our system, either as a baseline for future research, or for experimenting alongside human fact-checkers.

\subsection{Future works}
\begin{enumerate}
    \item Integrating a search API for use in \review{real-world applications} 
    \item Re-examine the Likert-scale rating (section~\ref{likert}) to establish a more appropriate and fine-grained means of tokenizing the label probabilities
    \item Generating evidence in the form of declarative sentences rather than Question-Answer pairs should be explored to see if it leads for better or worse fact-checking performance
    \item RAG-tuned LLMs such as those introduced in~\cite{menick2022teachinglanguagemodelssupport} could be explored to see if they offer a more reliable source citing
\end{enumerate}

\section*{Limitations}
The evaluation of our fact-checking pipeline is limited to the English language and the \averitec{} dataset~\cite{averitec2024}. This is a severe limitation as the pipeline when deployed in a real-world application, would encounter other languages and forms of claims not covered by the used dataset.

Another limitation is that we are using a large language model. Because of that, future usage is limited to using an API of a provider of LLMs or having access to a large amount of computational resources, which comes at significant costs. Using APIs also brings the disadvantage of sending data to a third party, which might be a security risk in some critical applications. LLM usage also has an undeniable environmental impact because of the vast amount of electricity and resources used.

The reliability of the generated text is a limitation that is often linked to LLMs. LLMs sometimes hallucinate (in our case, it would mean using sources other than those given in the system prompt), and they can be biased based on their extensive training data. Moreover, because of the dataset size, it is impossible to validate each output of the LLM, and thus, we are not able to 100\% guarantee the quality of the results.

\section*{Ethics statement}
It is essential to note that our pipeline is not a real fact-checker that could do a human job but rather a study of future possibilities in automatic fact-checking and a showcase of the current capabilities of state-of-the-art language models. The pipeline in its current state should only be used with human supervision because of the potential biases and errors that could harm the consumers of the output information or persons mentioned in the claims. The pipeline could be misused to spread misinformation by directly using misinformation sources or by intentionally modifying the pipeline in a way that will generate wrong outputs.

Another important statement is that our pipeline was in its current form explicitly built for the \averitec{} shared task, and thus, the evaluation results reflect the bias of the annotators. For more information, see the relevant section of the original paper~\cite{averitec2024}.

The carbon costs of the training and running of our pipeline are considerable and should be taken into account given the urgency of climate change. At the time of deployment, the pipeline should be run on the smallest possible model that can still provide reliable results, and the latest hardware and software optimisations should be used to minimise the carbon footprint.

\section*{Acknowledgements}
We would like to thank Bryce Aaron from UNC for exploring the problems of search query generation and pinpointing claims of underrepresented labels using numerical methods that did not make it into our final pipeline but gave us a frame for comparison. 

This research was co-financed with state support from the Technology Agency of the Czech Republic and the Ministry of Industry and Trade of the Czech Republic under the TREND Programme, project FW10010200.
The access to the computational infrastructure of the OP VVV funded project CZ.02.1.01/0.0/0.0/16\_019/0000765 ``Research Center for Informatics'' is also gratefully acknowledged.
We would like to thank to \mbox{OpenAI} for providing free credit for their paid API via Researcher Access Program\footnote{\review{\url{https://openai.com/form/researcher-access-program/}}}.


\appendix


\lstset{
    language={},
    basicstyle=\ttfamily\footnotesize\linespread{0.9}, 
    keywordstyle=\color{blue}\bfseries,
    commentstyle=\color{green!50!black}\itshape,
    stringstyle=\color{orange},
    numberstyle=\tiny\color{gray},
    numbers=none, 
    stepnumber=1, 
    numbersep=5pt, 
    tabsize=4, 
    showstringspaces=false, 
    breaklines=true, 
    breakatwhitespace=true,
    frame=lines, 
    captionpos=b, 
    breakindent=1em,
}
\begin{figure*}
    \section{System prompt}
    \label{appendix_sec:system_prompt}
    \begin{lstlisting}[breaklines=true, language={}, frame=single, caption={System prompt for the LLMs, \averitec{} claim is to be entered into the user prompt. Three dots represent omitted repeating parts of the prompt.}, label={lst:llm_system_prompt}]
You are a professional fact checker, formulate up to 10 questions that cover all the facts needed to validate whether the factual statement (in User message) is true, false, uncertain or a matter of opinion. Each question has one of four answer types: Boolean, Extractive, Abstractive and Unanswerable using the provided sources.
After formulating Your questions and their answers using the provided sources, You evaluate the possible veracity verdicts (Supported claim, Refuted claim, Not enough evidence, or Conflicting evidence/Cherrypicking) given your claim and evidence on a Likert scale (1 - Strongly disagree, 2 - Disagree, 3 - Neutral, 4 - Agree, 5 - Strongly agree). Ultimately, you note the single likeliest veracity verdict according to your best knowledge.
The facts must be coming from these sources, please refer them using assigned IDs:
---
## Source ID: 1 [url]
[context before]
[page content]
[context after]
...

---
## Output formatting
Please, you MUST only print the output in the following output format:
```json
{
 "questions":
     [
         {"question": "<Your first question>", "answer": "<The answer to the Your first question>", "source": "<Single numeric source ID backing the answer for Your first question>", "answer_type":"<The type of first answer>"},
         {"question": "<Your second question>", "answer": "<The answer to the Your second question>", "source": "<Single numeric Source ID backing the answer for Your second question>", "answer_type":"<The type of second answer>"}
     ],
 "claim_veracity": {
     "Supported": "<Likert-scale rating of how much You agree with the 'Supported' veracity classification>",
     "Refuted": "<Likert-scale rating of how much You agree with the 'Refuted' veracity classification>",
     "Not Enough Evidence": "<Likert-scale rating of how much You agree with the 'Not Enough Evidence' veracity classification>",
     "Conflicting Evidence/Cherrypicking": "<Likert-scale rating of how much You agree with the 'Conflicting Evidence/Cherrypicking' veracity classification>"
 },
 "veracity_verdict": "<The suggested veracity classification for the claim>"
}
```
---
## Few-shot learning
You have access to the following few-shot learning examples for questions and answers.:

### Question examples for claim "{example["claim"]}" (verdict {example["gold_label"]})
"question": "{question}", "answer": "{answer}", "answer_type": "{answer_type}"
...
    \end{lstlisting}
\end{figure*}

\section{Examples of errors}
\label{appendix_sec:errors}
\lstset{
    basicstyle=\ttfamily\footnotesize\linespread{0.9}, 
    numbers=none, 
    stepnumber=1, 
    numbersep=5pt, 
    tabsize=4, 
    showstringspaces=false, 
    breaklines=true, 
    breakatwhitespace=true,
    frame=lines, 
    captionpos=b, 
    breakindent=1em,
    frame=single
}

\begin{lstlisting}[caption={Example of a claim where our pipeline uses newspaper sources instead of official government sources.}, label={lst:gov_error}, breaklines=true, breakatwhitespace=false, columns=fullflexible]
Claim 479: Donald Trump said "When the anarchists started ripping down our statues and monuments, right outside, I signed an order immediately, 10 years in prison."

gold evidence example: 
    question: What was the law signed by Trump regarding damaging federal property?
    answer: Trump signed an executive order that authorizes a penalty of up to 10 years in prison for damaging federal property. It does not say that it will automatically be a 10 year prison sentence.,
    source_url: https://web.archive.org/web/20210224033536/https://www.whitehouse.gov/presidential-actions/executive-order-protecting-american-monuments-memorials-statues-combating-recent-criminal-violence/

pipeline evidence example: 
    question: Did Trump sign an order related to vandalism of statues and monuments?, 
    answer: Yes, Trump signed an executive order to prosecute those who damage national monuments, making it a punishable offense with up to 10 years in jail.,
    url: https://m.economictimes.com/news/international/world-news/trump-makes-vandalising-national-monuments-punishable-offence-with-up-to-10-yrs-jail/articleshow/76658610.cms
    
\end{lstlisting}

\begin{lstlisting}[caption={Example of a claim and questions showing that the last tends to be unrelated or redundant to fact-checking of the claim.}, label={lst:unrelated_questions}, breaklines=true, breakatwhitespace=false, columns=fullflexible]
Claim 295: Trump campaign asked Joe Biden to release a list of potential Supreme Court picks only after Ginsburg's passing
question 1: Did Joe Biden claim that the Trump campaign asked him to release a list of potential Supreme Court picks only after Ginsburg's passing?
question 2: Did the Trump campaign ask Joe Biden to release a list of potential Supreme Court picks before Ginsburg's passing?
question 3: When did Trump release his latest list of potential Supreme Court nominees?
question 4: Did Trump personally demand that Biden release a list of potential Supreme Court nominees before Ginsburg's death?
question 5: What did Trump say about Biden releasing a list of potential Supreme Court nominees during the Republican National Convention?
question 6: Did the Trump campaign issue a statement on September 17, 2020, regarding Biden releasing a list of potential Supreme Court nominees?
question 7: What did the Trump campaign's statement on September 9, 2020, say about Biden releasing a list of potential Supreme Court nominees?
question 8: Did Biden indicate in June 2020 that he might release a list of potential Supreme Court picks?
quetion 9: What reason did Biden give for not releasing a list of potential Supreme Court nominees?,
question 10: Did Biden pledge to nominate a Black woman to the Supreme Court?
    
\end{lstlisting}

\begin{lstlisting}[caption={Example of a claim where our pipeline did not exactly extract the answer.}, label={lst:extractive_error}, breaklines=true, breakatwhitespace=false, columns=fullflexible]
Claim #155 - Trump said 'there were fine people on both side' in far-right protests.
answer: "You had some very bad people in that group, but you also had people that were very fine people, on both sides.", 
answer_type: Extractive
url: https://www.theatlantic.com/politics/archive/2017/08/trump-defends-white-nationalist-protesters-some-very-fine-people-on-both-sides/537012/
scraped text: ... "You also had some very fine people on both sides," he said. The Unite the Right rally that sparked the violence in Charlottesville featured several leading names in the white-nationalist alt-right movement, and also attracted people displaying Nazi symbols. ...
\end{lstlisting}

\begin{lstlisting}[caption={An example of a claim where the evidence consists mainly of evidence from PolitiFact and Factcheck.org fact-checking articles leading to different predicted label than in the gold dataset}, label={lst:polifact_leakage}, breaklines=true, breakatwhitespace=false, columns=fullflexible]
Claim #483 - Donald Trump said "We have spent nearly $2.5 trillion on completely rebuilding our military, which was very badly depleted when I took office."
Gold Label: Not Enough Evidence
Predicted Label: Refuted
pipeline evidence example: 
    question: What is the total defense budget for the last four fiscal years under Trump?
    url: https://www.politifact.com/factchecks/2020/jan/10/donald-trump/trump-exaggerates-spending-us-military-rebuild/
    question: Did Trump spend $2.5 trillion specifically on rebuilding the military?
    url: https://www.factcheck.org/2020/07/trumps-false-military-equipment-claim/
    ...
\end{lstlisting}
\pagebreak
\begin{lstlisting}[caption={An example of a claims which differs in length.}, label={lst:different_lengths}, breaklines=true, breakatwhitespace=false, columns=fullflexible]
Claim #0 - In a letter to Steve Jobs, Sean Connery refused to appear in an apple commercial.
Gold Evidence:
    question: Where was the claim first published
    answer: It was first published on Sccopertino
    question: What kind of website is Scoopertino
    answer: Scoopertino is an imaginary news organization devoted to ferreting out the most relevant stories in the world of Apple, whether or not they actually occurred - says their about page

Claim #315 - The fastest Supreme Court justice ever confirmed in the U.S. was 47 days.
Gold Evidence:
    question: What is the quickest time a Supreme Court justice nomination has been confirmed in the United States?
    answer: John Paul Stevens waited the fewest number of days (19)-followed by the most recent nominee to the Court, Amy Coney Barrett (27).61
    question: What is the average number of days between a nomination for a Supreme Court justice and the final Senate vote?
    answer: Overall, the average number of days from nomination to final Senate vote is 68.2 days (or approximately 2.2 months), while the median is 69.0 days.62 Of the 9 Justices currently serving on the Court, the average number of days from nomination to final Senate vote is 72.1 days (or approximately 2.4 months), while the median is 73.0 days. Among the current Justices, Amy Coney Barrett waited the fewest number of days from nomination to confirmation (27), while Clarence Thomas waited the greatest number of days (99).
\end{lstlisting}

\end{document}